\documentclass[sigplan,10pt]{acmart}

\renewcommand\footnotetextcopyrightpermission[1]{}
\renewcommand{\paragraph}[1]{\medskip\noindent\textbf{#1}}

\usepackage{xspace}

\definecolor{commentgreen}{rgb}{0, 0.5, 0}
\usepackage[
  vlined,
  linesnumbered,
]{algorithm2e}
\DontPrintSemicolon{}
\SetKwComment{Comment}{$\triangleright$\ }{}

\SetCommentSty{mycommfont}
\newcommand{\algrule}[1][.7pt]{\par\vskip.5\baselineskip\hrule height #1\par\vskip.5\baselineskip}
\let\oldnl\nl
\newcommand{\nonl}{\renewcommand{\nl}{\let\nl\oldnl}}

\usepackage{float}
\usepackage[english]{babel}
\usepackage{subfig}
\usepackage{colortbl}
\usepackage{pifont}
\usepackage[inline]{enumitem}
\usepackage{multirow}
\usepackage{fancybox} %
\usepackage[normalem]{ulem}
\usepackage{comment}
\usepackage{booktabs}

\usepackage{tikz}
\usetikzlibrary{automata, positioning, arrows}

\newcommand{\rqt}[1]{\kern0.1em{\setlength{\fboxsep}{1pt}\setlength{\fboxrule}{0.4pt}\fbox{\makebox[0.5em][c]{\rule{0pt}{0.5em}\small#1}}}\kern0.2em}

\newcommand{\eat}[1]{}

\theoremstyle{definition}

\newtheorem*{defi*}{Definition}

\usepackage{listings}
\usepackage{courier}
\newenvironment{denseitemize}{
	\begin{itemize}[topsep=2pt, partopsep=0pt, leftmargin=1.5em]
		\setlength{\itemsep}{2pt}
		\setlength{\parskip}{0pt}
		\setlength{\parsep}{0pt}
	}{\end{itemize}}

\def\noeditingmarks{}
\usepackage{peanutgallery}

\begin{document}

  \title[Cornfigurator: Automated Planning for Any-to-Any Multimodal Model Serving]{\LARGE Cornfigurator: Automated Planning for Any-to-Any Multimodal Model Serving}

  \author{Jeff J. Ma$^{\text{1},\ast}$\enskip Jae-Won Chung$^{\text{1},\ast}$\enskip Jisang Ahn$^{\text{1}}$\enskip Yizhuo Liang$^{\text{2}}$\enskip Runyu Lu$^{\text{1}}$\enskip \\ Akshay Jajoo$^{\text{3}}$\enskip Myungjin Lee$^{\text{3}}$\enskip Mosharaf Chowdhury$^{\text{1}}$}
  \affiliation{\vspace{1mm} $^{\text{1}}$University of Michigan \enskip $^{\text{2}}$University of Southern California \enskip $^{\text{3}}$Cisco Research \country{}}
  \renewcommand{\shortauthors}{Ma, Chung, et al.}

  \begin{abstract}
	{\renewcommand{\thefootnote}{\fnsymbol{footnote}}\footnotetext[1]{Equal contribution.}}
	Any-to-Any models are an emerging class of multimodal models that accept combinations of text and multimodal data as input and generate them as output, introducing heterogeneous computation paths and component scaling characteristics.
There are existing mechanisms for deploying Any-to-Any models---or special cases of them---for inference serving, but they either require manual effort and expertise to tune, or do not generalize to generic Any-to-Any models.

We present Cornfigurator, the first deployment planner for generic Any-to-Any model inference serving.
The goal of Cornfigurator is to maximize the overall \emph{goodput} of serving the model, defined as the throughput of requests meeting their latency targets.
To do so, based on model and workload characteristics, Cornfigurator explores the full spectrum of deployment strategies, from colocation to disaggregation and mixing different strategies. 
Cornfigurator performs coarse-to-fine statistical evaluation to efficiently navigate the large space of candidate plans.
Plans generated by Cornfigurator either match or deliver 1.12$\times$--6.32$\times$ higher goodput compared to existing systems and expert-tuned deployment plans.
Cornfigurator is open-source available on Github\footnote{\url{https://github.com/cornserve-ai/cornfigurator}}.

  \end{abstract}

	\maketitle

	\section{Introduction}\label{sec:intro}

\begin{figure}[t]
  \centering
  \subfloat[Multimodal input, text output (MLLM)]{
    \includegraphics[width=0.35\textwidth]{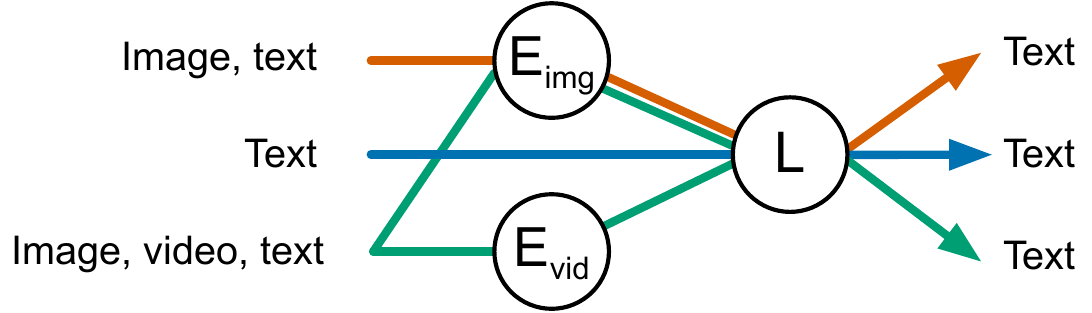}
    \label{fig:intro-a2a-mllm}
  }

  \subfloat[Multimodal input, multimodal output]{
    \includegraphics[width=0.40\textwidth]{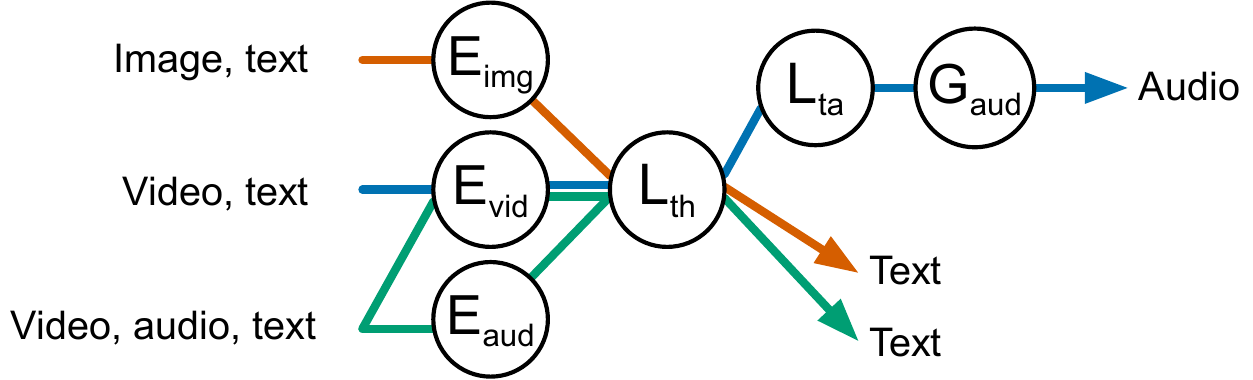}
    \label{fig:intro-a2a-omni}
  }
  \caption{
    Requests invoking (a) a multimodal input model (InternVL~\cite{internvl3-arxiv25}) and (b) a multimodal input and output model (Qwen Omni~\cite{qwen2.5-omni-arxiv25,qwen3-omni-arxiv25,qwen3.5-omni-blog}).
    Different requests invoke different components of the model in different paths.
    $E$ stands for Encoder, $L$ for LLM, and $G$ for Generator.
    $L_{\text{th}}$ and $L_{\text{ta}}$ stand for thinker and talker LLMs, respectively.}
  \label{fig:intro-a2a-models}
\end{figure}

Going beyond text-only Large Language Models (LLMs), we see a rapid proliferation of multimodal models that process and generate not just text, but also images, video, and audio.
These \emph{Any-to-Any} models, with over 11,000 variants on Hugging Face as of March 2026~\cite{hf-any-to-any-models}, can
(1) understand multimodal inputs and/or
(2) generate multimodal outputs alongside text.
For instance, Multimodal LLMs (MLLMs) like Qwen VL~\cite{qwen2.5-vl-arxiv25,qwen3-vl-arxiv25} and InternVL~\cite{internvl3-arxiv25} process multimodal inputs and generate text (Figure~\ref{fig:intro-a2a-mllm});
Qwen Image~\cite{qwen-image-arxiv25,qwen-image-2-blog} and GLM Image~\cite{glm-image-blog} produce images using diffusion from text embedded by an LLM;
LTX-2~\cite{ltx2-arxiv26} generates video and audio;
DeepSeek Janus~\cite{janus-arxiv24,janus-pro-arxiv25} understands and generates both text and images;
and Qwen Omni~\cite{qwen2.5-omni-arxiv25,qwen3-omni-arxiv25,qwen3.5-omni-blog} accepts combinations of text, image, video, and audio as input and generates text and audio (Figure~\ref{fig:intro-a2a-omni}).
In essence, text-only LLMs or diffusion models that generate images and videos are \emph{special cases} of Any-to-Any multimodal models.

The computations of Any-to-Any (A2A) models are defined as a \emph{graph of heterogeneous components} that handle different modalities: multimodal encoders, autoregressive components like LLMs, and multimodal generators.
The execution of A2A models is distinguished from that of traditional special cases by two new types of heterogeneity (\S\ref{sec:motivation-a2a-models}).
First, different \emph{request types} traverse different computation paths through the graph.
Figure~\ref{fig:intro-a2a-models} illustrates this with MLLMs and Qwen Omni as examples.
Particularly, in Qwen Omni (Figure~\ref{fig:intro-a2a-omni}), image, video, and audio encoders feed embeddings into a \emph{thinker LLM} for text generation; then, if audio output was requested by the user, the thinker's output is further passed to a \emph{talker LLM} and then a vocoder, producing audio waveforms.
Requests with different input and output modalities therefore invoke different subsets of components, leading to uneven per-component request rates.
Second, different components have vastly different resource requirements and computational characteristics.
As we show in Section~\ref{sec:motivation-a2a-models}, in Qwen 3 Omni, for instance, the thinker LLM achieves nearly 30$\times$ higher request throughput than the talker LLM on A100.
Without well-balanced resource allocation to heterogeneous components, the overall throughput of inference serving will be bottlenecked by the slowest component.

Existing works have handled component heterogeneity in large model serving by \emph{disaggregating} computation into separate \emph{executors}~\cite{vllm-omni-github,sglang-omni-github,modserve-socc26,epd-icml25}.
On the one hand, systems like vLLM-Omni~\cite{vllm-omni-github} and SGLang-Omni~\cite{sglang-omni-github} provide mechanisms to disaggregate and run inference of generic A2A models, but require human experts to manually search for a good deployment plan.
On the other hand, systems like ModServe~\cite{modserve-socc26} and EPD~\cite{epd-icml25} are designed for special cases of A2A models (e.g., MLLMs) and do not generalize to generic A2A models.
Neither category of systems provides an automated planner for generic A2A models, and building such a planner is non-trivial; the best strategy can be complex and model-dependent, and finding the best strategy requires navigating a large and complex search space (\S\ref{sec:motivation-deployment}).

To fill this gap, we build Cornfigurator, an automated deployment planner for generic A2A model inference serving (\S\ref{sec:overview}).
Cornfigurator is designed to make the right colocation and disaggregation decisions for A2A models based on model and workload characteristics, instead of prescribing fixed strategies based on model architecture.
The key insight of Cornfigurator is that we should reason about \emph{each request type}, instead of lumping all requests together.
This is because different request types invoke different amounts of computation and serve different purposes within an application, even though they are served by a \emph{shared} model.
For instance, an audio response and a text response may have different latency expectations.
Thus, Cornfigurator's optimization objective is to maximize the throughput of each request type constrained by each type's own latency target, or in other words, to maximize the \emph{goodput} of each request type.

To do so, Cornfigurator's planning algorithm (\S\ref{sec:design}) systematically explores colocation and disaggregation combinations for a given model, rather than prescribing a fixed strategy.
The planner enumerates \emph{logical subplans} (graph topologies) that may each \emph{specialize} for different subsets of request types, merges subplans that share nodes into \emph{compound subplans}, and composes them into \emph{logical plans}.
Logical plans are annotated with per-node executor configurations and routing probabilities to produce \emph{physical plans}, which are concrete specifications of how to deploy and run the model.
Then, the planner evaluates the per-request-type goodput of each physical plan through coarse-to-fine statistical evaluation: network flow for throughput ceiling, Monte Carlo sampling for latency, and a request-level simulator for accurate serving dynamics modeling, with pruning at each stage.

Cornfigurator is runtime-agnostic by design.
We implement and evaluate it on top of Cornserve~\cite{cornserve-github, cornserve-cais26}, a distributed serving runtime for generic A2A models (\S\ref{sec:implementation}).
On a variety of recent A2A models including Qwen 3 Omni~\cite{qwen3-omni-arxiv25}, Qwen 3 VL~\cite{qwen3-vl-arxiv25}, InternVL 3~\cite{internvl3-arxiv25}, and Qwen Image~\cite{qwen-image-arxiv25}, Cornfigurator's plans either match or deliver 1.12$\times$--6.32$\times$ higher goodput compared to plans used by existing systems (\S\ref{sec:evaluation}).

To summarize, our contributions are as follows:
\begin{denseitemize}
  \item We identify a gap in automated deployment planning for generic Any-to-Any model serving.
  \item We present Cornfigurator, an automated planner for generic Any-to-Any models that maximizes goodput by reasoning about each request type and navigating deployment strategies and resource allocations.
  \item We evaluate Cornfigurator on a variety of recent Any-to-Any models and show that its plans match or deliver higher goodput compared to plans in existing systems.
\end{denseitemize}

	\section{Background and Motivation}\label{sec:motivation}

\looseness=-1
We provide background on Any-to-Any multimodal models (\S\ref{sec:motivation-a2a-models}), and discuss the deployment configuration space that motivates the need for an automated planner (\S\ref{sec:motivation-deployment}).

\subsection{Any-to-Any Multimodal Models}\label{sec:motivation-a2a-models}

\begin{table}[t]
  \footnotesize
  \centering
  \begin{tabular}{lll}
    \toprule
    \textbf{Model} & \textbf{Input} & \textbf{Output} \\
    \midrule
    Qwen 2.5/3/3.5 Omni~\cite{qwen2.5-omni-arxiv25,qwen3-omni-arxiv25,qwen3.5-omni-blog} & T, I, V, A & T, A \\
    Qwen 2.5/3 VL~\cite{qwen2.5-vl-arxiv25,qwen3-vl-arxiv25}, InternVL 3~\cite{internvl3-arxiv25} & T, I, V & T \\
    DeepSeek Janus~\cite{janus-arxiv24,janus-pro-arxiv25} & T, I & T, I \\
    LTX-2~\cite{ltx2-arxiv26} & T, I & V, A \\
    Qwen Image~\cite{qwen-image-arxiv25}, GLM Image~\cite{glm-image-blog} & T & I \\
    \bottomrule
  \end{tabular}
  \vspace{0.5em}
  \caption{Input and output modalities of recent Any-to-Any multimodal models. Modalities (\textbf{T}ext, \textbf{I}mage, \textbf{V}ideo, \textbf{A}udio) supported by a model can vary significantly.}\label{tab:background-a2a-models}
\end{table}

\begin{table}[t]
  \vspace{-0.5em}
  \setlength{\tabcolsep}{3.3pt}
  \footnotesize
  \centering
  \begin{tabular}{lrrrrrr}
    \toprule
    \begin{tabular}{@{}l@{}}\phantom{\textbf{Image}}\\\textbf{Model}\end{tabular} & \begin{tabular}{@{}r@{}}\textbf{Image}\\\textbf{input}\end{tabular} & \begin{tabular}{@{}r@{}}\textbf{Video}\\\textbf{input}\end{tabular} & \begin{tabular}{@{}r@{}}\textbf{Audio}\\\textbf{input}\end{tabular} & \begin{tabular}{@{}r@{}}\textbf{Text}\\\textbf{in/out}\end{tabular} & \begin{tabular}{@{}r@{}}\textbf{Audio}\\\textbf{output}\end{tabular} & \begin{tabular}{@{}r@{}}\textbf{Image}\\\textbf{output}\end{tabular}\\
    \midrule
    Qwen 3 Omni   &  5.43 & 2.93 & 21.43 &  2.15 & 0.12 &      \\
    Qwen 2.5 Omni & 15.64 & 1.28 & 34.04 &  1.09 & 0.28 &      \\
    Qwen 3 VL     &  8.95 & 0.97 &       &  1.56 &      &      \\
    Qwen 2.5 VL   & 12.04 & 0.89 &       &  1.63 &      &      \\
    InternVL 3    &  1.13 & 0.74 &       &  0.59 &      &      \\
    Qwen-Image    &       &      &       & 15.67 &      & 0.20 \\
    \bottomrule
  \end{tabular}
  \vspace{0.5em}
  \caption{
    Per-component throughput (requests/s) of various Any-to-Any models on A100-80GB GPU.
    Empty cells mean the model does not have that component.}\label{tab:motivation-tput-differences}
\end{table}

\begin{figure*}
  \begin{center}
    \includegraphics[width=0.90\textwidth]{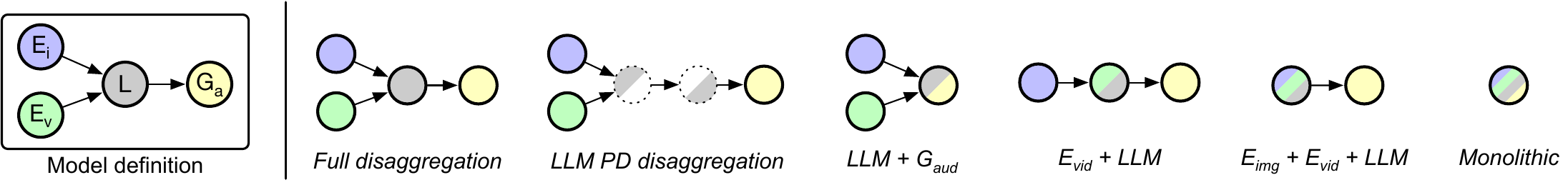}
  \end{center}
  \vspace{-1em}
  \caption{
    Graph-level deployment strategies for a four-component Any-to-Any model.
    The leftmost graph shows the model's component graph definition: (E$_\textrm{img}$, E$_\textrm{vid}$) $\to$ LLM $\to$ G$_\textrm{aud}$.
    Graphs to the right show various example graph-level deployment strategies, which differ in how they group or split components into nodes (circles).
    Strategies shown are non-exhaustive.
  }\label{fig:motivation-graph-deployment}
\end{figure*}

\looseness=-1
Recent models embrace multimodality at their core, naturally so because our world is full of multimodal information and interactions.
This gave rise to a new class of models called \emph{Any-to-Any} (A2A) multimodal models.
As Figure~\ref{fig:intro-a2a-models} shows, A2A model computations form a graph of heterogeneous components.
In essence, traditional text-only Large Language Models (LLMs) and Diffusion Transformer (DiT) models are \emph{special cases} of A2A models where all requests traverse the same path through a linear pipeline of components.

When it comes to deployment and inference, the graph-based computation structure of A2A models introduces new sources of heterogeneity.
Table~\ref{tab:background-a2a-models} summarizes the input and output modalities supported by recent A2A models.
Requests with different combinations of input and output modalities invoke different subgraphs---we call each combination with a unique subgraph a \emph{request type}.
An incoming inference workload contains a mix of request types, leading to each component experiencing a different request rate.
Furthermore, different components have vastly different resource requirements and computational characteristics.
Table~\ref{tab:motivation-tput-differences} shows the per-component throughput of recent A2A models on an A100-80GB GPU.\footnote{Requests are from ServeGen~\cite{servegen-nsdi26}. We use executor default configurations, except for LLMs larger than 30B for which we use tensor parallel degree 2.}
Component throughput can differ by orders of magnitude within the same model: in Qwen 3 Omni, the throughput difference between audio input and output components is nearly 200$\times$.
Together with request type heterogeneity, component heterogeneity leads to even more uneven per-component load.

\subsection{Deployment Strategies and Configurations}\label{sec:motivation-deployment}

The heterogeneity of A2A models motivates specialized deployment strategies---namely, grouping components into \emph{nodes} (a colocated set of components; graph-level), and configuring each node's execution (executor-level).
These decisions are associated with complex tradeoffs and lack any silver bullet strategy optimal for all models and workloads.

\paragraph{Graph-level decisions.}
At the graph-level, the main decision is how to group components into graph nodes.
Prior works focused on MLLMs, a special case of A2A models that consist of an LLM and one or more multimodal encoders.
The simplest strategy deploys the entire model in a single \emph{monolithic} executor (e.g., vLLM~\cite{pagedattention-sosp23}), but this couples the execution and scaling of all components, making the slowest component the bottleneck for the entire model.
\emph{Disaggregated} strategies decouple the execution and scaling of different components by placing them on separate devices.
For MLLMs, we may disaggregate just the encoders (e.g., ModServe~\cite{modserve-socc26}) or further disaggregate the LLM into separate Prefill and Decode phases (e.g., EPD disaggregation~\cite{epd-icml25}).
Going further, systems like vLLM-Omni~\cite{vllm-omni-github} and SGLang-Omni~\cite{sglang-omni-github} provide generalized component-wise and/or phase-wise disaggregation for A2A models.
Figure~\ref{fig:motivation-graph-deployment} shows graph-level strategies for an example four-component A2A model.

\paragraph{Executor-level decisions.}
After we make graph-level decisions, we know which components are grouped together in an executor.
Each executor exposes its own configuration knobs that control its throughput, latency, and GPU resource consumption.
For instance, nearly all executors can be configured with a maximum \emph{batch size} and a \emph{parallelism degree} with different parallelism strategies available depending on the executor and component (e.g., tensor/expert parallelism for LLMs~\cite{pagedattention-sosp23}, sequence parallelism for DiTs~\cite{tetriserve-asplos26}).

\begin{figure}[t]
  \centering
  \vspace{-1.0em}
  \subfloat[Throughput (req/s)]{%
    \includegraphics[height=4.0cm]{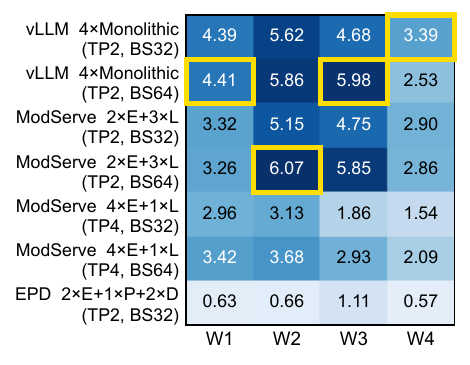}\label{fig:motivation-heatmap-tput}
  }%
  \subfloat[p90 latency (s)]{%
    \includegraphics[height=4.0cm]{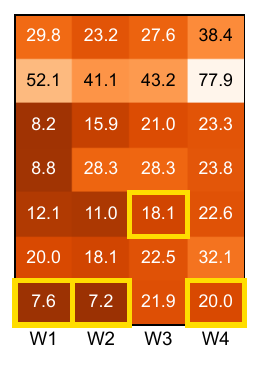}\label{fig:motivation-heatmap-lat}
  }%
  \vspace{-0.7em}
  \caption{
    InternVL 3~\cite{internvl3-arxiv25} (a) throughput and (b) p90 latency under different deployment strategies and workloads\protect\footnotemark{} on $8\times$ A100-80GB GPUs.
    Each row uses a graph-level strategy from prior work; numbers before \textbf{E}/\textbf{L}/\textbf{P}/\textbf{D} indicate executor instances.
    LLM (\textbf{L}/\textbf{D}) tensor parallel degree (TP) and batch size (BS) are shown in row labels; disaggregated encoders (\textbf{E}) and prefill (\textbf{P}) always use TP1 BS1 and TP2 BS1, respectively.
  }
  \label{fig:motivation-heatmap}
\end{figure}

\footnotetext{Workloads are specified as (input length, output length, image resolution, number of images per request): W1=(100, 100, 1920$\times$1080, 1), W2=(1000, 100, 896$\times$896, 1), W3=(100, 300, 896$\times$896, 1), W4=(100, 300, 896$\times$896, 2).}

\paragraph{No silver bullet.}
Disaggregation, while proposed by many prior works, is not universally beneficial.
Placing a component on its own GPU lets it scale independently, but it also consumes GPU resources that could otherwise be used by other components.
For instance, disaggregating a multimodal encoder from the LLM allows independent scaling, but the GPUs allocated to the encoder can no longer store the KV cache for the LLM, which may reduce the LLM's throughput.
At the executor-level, increasing parallelism degree reduces per-request latency by distributing compute across more GPUs, but may hurt throughput (and eventually latency) due to communication overhead.
Similarly, increasing batch size improves throughput until the GPU saturates, but it may hurt latency~\cite{andes-arxiv24}.
Navigating these tradeoffs and determining graph- and executor-level configuration knobs is highly dependent on the workload and model architecture, as shown by prior work on LLM serving~\cite{distserve-osdi24,sarathiserve-osdi24}, DiT serving~\cite{tetriserve-asplos26}, and configuration tuning~\cite{aiconfigurator-arxiv26,nvidia-dynamo}.
Figure~\ref{fig:motivation-heatmap} quantifies this for InternVL 3 38B~\cite{internvl3-arxiv25}.
The heatmap shows the throughput and latency of different graph- and executor-level configurations under different workloads on $8\times$ A100-80GB GPUs.
Even for this relatively simple single-encoder MLLM, there is no silver bullet strategy; the best-throughput and best-latency configurations vary across workloads.

\looseness=-1
\paragraph{Automated planning.}
Navigating the vast configuration space of A2A model deployment (order of \emph{millions}; \S\ref{sec:evaluation-overhead}) to find the best strategy is non-trivial, and existing works are limited.
They either are designed for special cases of A2A models and prescribe fixed deployment strategies based on model architecture~\cite{modserve-socc26,epd-icml25}, or only provide mechanisms to disaggregate and run specific A2A models~\cite{vllm-omni-github,sglang-omni-github}.
Building an efficient automated planner that can search for good deployment plans for generic A2A models is the goal of this paper.

	\section{Cornfigurator Overview}\label{sec:overview}

Cornfigurator is a deployment planner for Any-to-Any model inference serving.
This section presents the planner's location and role in the serving stack (\S\ref{sec:overview-architecture}) and the planning objective (\S\ref{sec:overview-objective}).
Section~\ref{sec:design} describes the planning algorithm.

\begin{figure}[t]
  \centering
  \includegraphics[width=0.87\columnwidth]{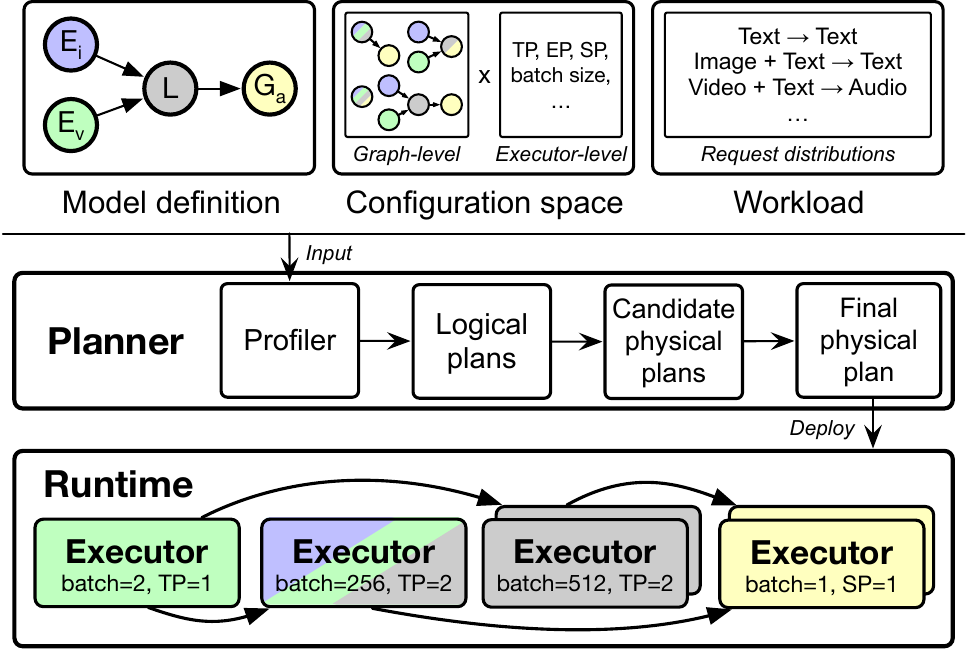}
  \caption{
    Planning and deployment with Cornfigurator.
    The planner takes as input the model definition, configuration space, and workload; the profiler benchmarks model components and the planner produces a physical plan.
    Cornfigurator assumes an existing serving runtime that deploys executors on GPUs and runs requests as planned.
  }\label{fig:overview-architecture}
\end{figure}

\subsection{Planning and Deployment Architecture}
\label{sec:overview-architecture}

Figure~\ref{fig:overview-architecture} shows how the planner fits into the serving stack.
The planner supplies an existing \emph{serving runtime} with a \emph{physical plan} derived from the following inputs:
\begin{denseitemize}
  \item \textbf{Model definition}: A directed acyclic graph whose nodes are model components and edges are data dependencies.
  \item \textbf{Configuration space}: The set of all executor types the runtime can instantiate, each specifying which model component(s) it handles, from which other executor types it can receive input, and available executor-level configurations. This determines the feasible graph-level and executor-level configurations (\S\ref{sec:motivation-deployment}).
  \item \textbf{Workload}: A representative set of requests that follows the expected distribution of request types $T$, with per-type fractions $\pi_t$ ($\sum_{t \in T} \pi_t = 1$).
  \item \textbf{Parameters}: Number of homogeneous GPUs available $N$ and latency targets $L_t$ for each request type.
\end{denseitemize}

Given these inputs, the \emph{profiler} (\S\ref{sec:impl-profiler}) benchmarks each model component under the target workload, recording the throughput and latency of each component on the target hardware.
Using these profiles, the \emph{planner} (\S\ref{sec:design}) produces a set of \emph{physical plans}, each specifying the nodes, the number of executors for each, their configurations, and routing probabilities when multiple paths are available for a request type.
The final physical plan (\S\ref{sec:overview-objective}) is deployed to the serving \emph{runtime}, which instantiates executors on the GPUs and serves requests according to the plan.

\subsection{Planning Objective}\label{sec:overview-objective}

\begin{figure}[t]
  \centering
  \includegraphics[width=0.93\columnwidth]{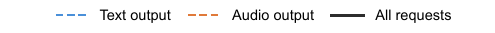}
  \includegraphics[width=0.68\columnwidth]{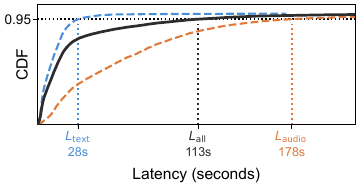}
  \vspace{-0.5em}
  \caption{
    Latency CDFs of text-output requests (blue), audio-output requests (orange), and all requests together (black).
    Audio-output requests require more computation as they must go through the talker LLM and vocoder.
    A global latency constraint $L_{\mathrm{all}}$ binds only on the heavier audio-output requests, while per-type latency targets $L_{\mathrm{text}}$ and $L_{\mathrm{audio}}$ ensure each request type is individually constrained.
  }\label{fig:overview-latency-cdf}
\end{figure}

The planner's goal is to maximize throughput subject to latency constraints.
A common formulation is to constrain a single global statistic of request latency, but this can be problematic for A2A models.
First, with input and/or output modality differences between request types, the application context within which the request's response is used may differ, leading to different latency expectations per type.
For instance, users may tolerate longer waits for videos than for images.
Furthermore, in generic A2A models, requests of different types take different paths through the model, leading to widely different computation costs and latency distributions.
Figure~\ref{fig:overview-latency-cdf} illustrates such a case, where an example A2A model has two request types: text output (blue; lighter) and audio output (orange; heavier).
When a single global latency target $L_{\mathrm{all}}$ (black) is used, it binds only on the heaviest request type (audio), while the lighter type (text) faces no effective constraint; nothing stops the planner from freely degrading the latency of text requests.\footnote{Lighter types with less compute requirement are especially vulnerable. Since their latency is far below the global threshold, they can absorb substantial degradation before ever pressuring the global constraint.}
Therefore, Cornfigurator instead imposes latency constraints on \emph{each request type independently}, ensuring that the latency of each type is constrained according to its own requirements and computation cost, and that the planner cannot arbitrarily degrade the latency of lighter request types.

\paragraph{Latency constraints.}
Cornfigurator allows user to set latency targets $L_t$ per type in any way that reflects their needs.
One natural default/starting point is to set $L_t$ proportionally to the computation cost of type $t$, so that a request requiring twice the computation gets twice the latency budget.
Appendix~\ref{sec:appendix-normalization} proves that this choice factors out scale differences between types, making the constraint equally tight for all types when their latency distributions have similar shapes.

\paragraph{Maximizing goodput.}
Per-type latency constraints motivate a per-type metric that captures both throughput and latency compliance.
We define the \emph{goodput} of request type $t$ for a physical plan as the throughput of type-$t$ requests that meet their latency target.
Goodput has desirable properties.
A plan that violates latency targets gets penalized even if its raw throughput is high, and when a plan cannot handle incoming load, queue build-up increases latency and goodput tends to zero.
For each candidate plan, the planner produces a vector of per-type goodput estimates.
The planner's goal is to find the plan that maximizes overall goodput, summed across all request types.

	\section{Planning Algorithm}\label{sec:design}

This section describes the planning algorithm.
We first walk through the process at a high level (\S\ref{sec:design-overview}), then describe the two big stages: enumerating valid plans (\S\ref{sec:design-enumeration}), and evaluating them to find the best one (\S\ref{sec:design-pruning}).
Finally, we discuss how the planner can adapt to deployment scenario changes (\S\ref{sec:design-adapting}).

\begin{figure*}[t]
  \centering
  \subfloat[Definitions]{%
    \raisebox{0.4cm}{\includegraphics[height=3.2cm]{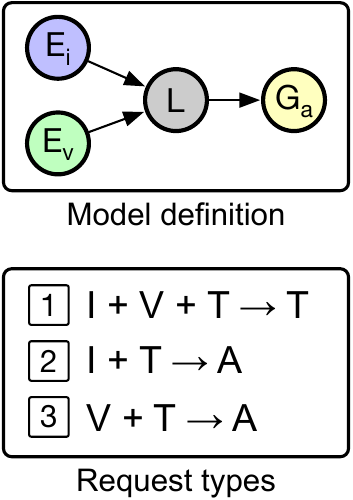}}\label{fig:pipeline-input}
  }%
  \hfill
  \subfloat[Simple and compound logical subplans]{%
    \raisebox{0.5cm}{\includegraphics[height=3cm]{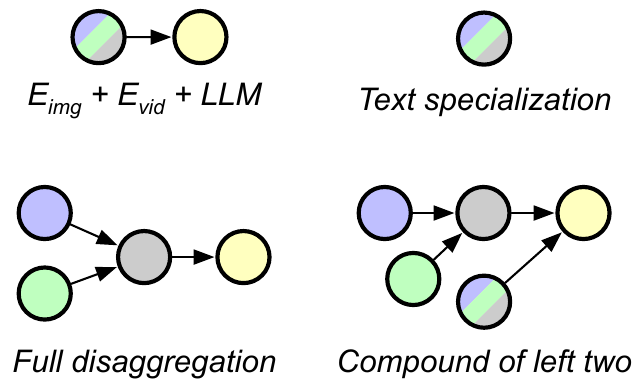}}\label{fig:pipeline-subplans}
  }%
  \hfill
  \subfloat[Logical plan]{%
    \raisebox{0.6cm}{\includegraphics[height=2.8cm]{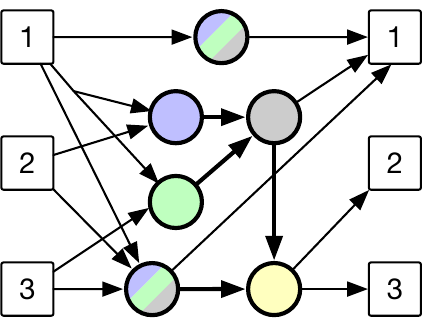}}\label{fig:pipeline-logical}
  }%
  \hfill
  \subfloat[Physical plan]{%
    \includegraphics[height=4cm]{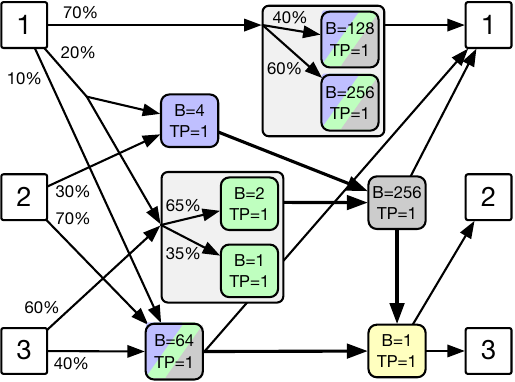}\label{fig:pipeline-physical}
  }%
  \caption{
    Running example for the plan enumeration phase.
    (a) Definition of a four-component model ((E$_\text{img}$, E$_\text{vid}$) $\to$ L $\to$ G$_\text{aud}$) with three request types (\rqt{1} I+V+T$\to$T, \rqt{2} I+T$\to$A, and \rqt{3} V+T$\to$A).
    (b) Example subplans (not exhaustive).
    Simple: E$_\text{img}$+E$_\text{vid}$+LLM with disaggregated G$_\text{aud}$, full disaggregation, and E$_\text{img}$+E$_\text{vid}$+LLM without G$_\text{aud}$ (specialized to type~\rqt{1}).
    Compound: merging the left two simple subplans sharing G$_\text{aud}$.
    (c) An example logical plan composing the compound subplan and the text specialization simple subplan into a supergraph.
    Audio-output types (\rqt{2} and \rqt{3}) must go to the compound subplan (only path with G$_\text{aud}$); type~\rqt{1} can go to either.
    Arrows that send text data to LLM executors are omitted for simplicity.
    (d) An example physical plan based on the logical plan in (c), allocating 8 GPUs, with per-node executors and their configurations, and per-type routing probabilities at the compound subplan entrypoint and across the two subplans.
  }\label{fig:design-planning-pipeline}
\end{figure*}

\subsection{Planning Overview}\label{sec:design-overview}

The goal of the planner is to explore diverse graph- and executor-level configurations and \emph{mixtures} of them to find smoother tradeoffs that strike the right balance for the given model, workload, and GPU budget.

Executor-level configurations can only be decided after the \emph{topology} of the graph is determined.
As such, the planner begins at the graph-level.
First, the planner enumerates \emph{simple logical subplans} by exploring colocation and disaggregation decisions on the model definition graph, and then merges simple subplans that share nodes into \emph{compound logical subplans} with internal parallelism and routing.
These simple and compound logical subplans are building blocks, and each may specialize for different subsets of request types.
Logical subplans are then composed into full \emph{logical plans} with parallel paths that collectively cover all request types.
These are then refined into \emph{physical plans} with concrete GPU allocations, configurations, and routing probabilities (\S\ref{sec:design-enumeration}).

Each physical plan is then evaluated for per-request-type goodput (\S\ref{sec:design-pruning}): a cheap statistical estimate prunes unpromising candidates, and a more accurate request-level simulator evaluates the survivors.
Finally, the physical plan that maximizes overall goodput is selected for deployment.

\subsection{Plan Enumeration}\label{sec:design-enumeration}

\begin{algorithm}[t]
  \algrule{}
  \KwIn{Model $G=(C,E)$, colocatable edges $E_c \subseteq E$\newline
    Config space $\mathcal{K}_v$ for each possible node $v$\newline
    Request types $T$, GPU budget $N$\newline
    Subplan merge limit $k_c$, composition limit $k_s$
  }
  \KwOut{Physical plans $\mathcal{P}$}
  \algrule{}

  \Comment{Simple subplans: colocation/disaggregation decisions}
  $\mathcal{L}_\text{simple} \leftarrow \emptyset$\;
  \Comment{Valid: covers $\ge 1$ request type, all components are used}
  \For{\textnormal{each valid subgraph} $G'=(C',E') \subseteq G$}{
    \Comment{One decision per colocatable edge in the subgraph}
    \For{\textnormal{each} $\mathbf{m} \in \{\textsc{Keep}, \textsc{Merge}\}^{|E' \cap E_c|}$\label{algline:keep-merge}}{
      $L \leftarrow$ \textnormal{fully disaggregated plan for} $G'$\;
      \For{\textnormal{each edge} $e=(c_i,c_j) \in E' \cap E_c$}{
        \lIf{$m_e = \textsc{Merge}$}{
          \textnormal{Merge} $c_i, c_j$ \textnormal{into} $c_{ij}$
        }
      }
      Add $L$ to $\mathcal{L}_\text{simple}$\;
    }
  }
  \Comment{Compound subplans: merge subplans with shared nodes}
  $\mathcal{L} \leftarrow \mathcal{L}_\text{simple}$\;
  \For{\textnormal{each subset} $\{L_1, \ldots, L_j\} \subseteq \mathcal{L}_0$, $2 \le j \le k_c$\label{algline:compound}}{
    \lIf{\textnormal{share nodes}}{
      Add overlay($L_1, \ldots, L_j$) to $\mathcal{L}$
    }
  }
  \Comment{Logical plans: compose subplans into supergraphs}
  $\mathcal{S} \leftarrow \emptyset$\;
  \For{\textnormal{each multiset} $S$ \textnormal{over} $\mathcal{L}$ \textnormal{where} $|S| \le k_s$\label{algline:compose}}{
    \lIf{\textnormal{covers all request types}}{Add $S$ to $\mathcal{S}$}
  }
  \Comment{Physical plans: allocate executors, configs, routing}
  $\mathcal{P} \leftarrow \emptyset$\;
  \For{\textnormal{each} $S \in \mathcal{S}$}{
    \Comment{Valid: uses $\le N$ GPUs, at least 1 executor per node}
    \For{\textnormal{each valid executor partition}\label{algline:executor-partition}}{
      \Comment{Valid: sums to 1.0 per request type}
      \For{\textnormal{each routing probability assignment}\label{algline:routing}}{
        Add $(S, \textnormal{executors}, \textnormal{routing})$ to $\mathcal{P}$\;
      }
    }
  }
  \Return{$\mathcal{P}$}
  \algrule{}
  \caption{Plan enumeration.}\label{algo:design-enumeration}
\end{algorithm}

Algorithm~\ref{algo:design-enumeration} summarizes the plan enumeration stage, and Figure~\ref{fig:design-planning-pipeline} provides a running example.
Given inputs (Figure~\ref{fig:pipeline-input}; \S\ref{sec:overview-architecture}), the planner enumerates candidates from logical subplans, to logical plans, and finally to physical plans.
Logical (sub)plans define empty \emph{variables} for GPU allocations, executor configurations, and routing probabilities.
Physical plans \emph{fill in} all such variables, producing complete specifications that can be deployed and served by a runtime.

\paragraph{Simple logical subplans.}
A \emph{logical subplan} is a directed acyclic graph, where each node handles one or more model components.
The planner first constructs \emph{simple} subplans by selecting a subgraph of the model definition that (1) covers at least one request type (not necessarily all types), and (2) has no unused components (i.e., every component in the subgraph is used by at least one covered request type).
Each pair of adjacent components connected by a \emph{colocatable} edge, as determined by the runtime, has a choice of whether to \textsc{Keep} as separate nodes or \textsc{Merge} into a single colocated node.
The Cartesian product of these binary decisions (Algorithm~\ref{algo:design-enumeration}, line~\ref{algline:keep-merge}) across all colocatable edges produces all valid node topologies for that subgraph---from fully disaggregated (all \textsc{Keep}) to monolithic (all \textsc{Merge}).

\paragraph{Compound logical subplans.}
The planner then generates \emph{compound} subplans (Algorithm~\ref{algo:design-enumeration}, line~\ref{algline:compound}) by merging up to $k_c$ simple subplans (default $k_c = 2$) that share one or more nodes.
Requests still follow one subplan end-to-end, and per-type routing probabilities at the entrypoint of the compound subplan determine how traffic is split between merged subplans.
Figure~\ref{fig:pipeline-subplans} shows example simple and compound subplans.

\paragraph{Constructing logical plans.}
A single logical subplan covers a subset of request types and occupies a portion of the GPU budget.
This design is intentional; it allows the planner to include subplans that are specialized to different request types.
For instance, one subplan may be optimized for text-output requests, while another handles audio-output requests (Figure~\ref{fig:pipeline-subplans}).
By having subplans that specialize for different subsets of request types, the planner can mix and match them as appropriate to produce smoother tradeoffs.
Given all subplans, the planner composes up to $k_s$ subplans (default $k_s = 2$) into a \emph{logical plan} (Algorithm~\ref{algo:design-enumeration}, line~\ref{algline:compose}): a supergraph with common entry and exit nodes, where each subplan forms an alternative parallel path.
A logical plan covers every request type with at least one path (Figure~\ref{fig:pipeline-logical}).

\paragraph{Constructing physical plans.}
A physical plan (Figure~\ref{fig:pipeline-physical}) is a logical plan annotated with concrete GPU allocations, executors and their configurations, and routing probabilities---it can be deployed to and run by a runtime.

First, for each node, the planner decides how many executors to run and what configuration (e.g., parallelism degree,\footnote{The concrete parallelism strategy depends on the component type: tensor parallelism (TP) for LLMs and Vision Transformers, expert parallelism (EP) for MoE models, and sequence parallelism (SP) for Diffusion Transformers.} batch size) each executor uses.
This enumeration is analogous to the coin change problem: each executor type is a ``coin'' whose denomination is its GPU cost, and allocating executors to nodes is equivalent to finding all ways to spend between 1 to $N$ GPUs.
Unlike standard coin change, however, coins are not fungible even when they have the same denomination: two executors that cost the same number of GPUs but differ in other configurations (e.g., batch size) are considered different.
The planner enumerates all physical plans (Algorithm~\ref{algo:design-enumeration}, line~\ref{algline:executor-partition}) satisfying:
\begin{equation}\label{eq:coin-change}
\begin{alignedat}{2}
  &\sum_{v \in S}\sum_{k \in \mathcal{K}_v} a_{v,k} n_{v,k} \le N &\qquad &\text{\small(Within GPU budget)} \\
  &\sum_{k \in \mathcal{K}_v} n_{v,k} \ge 1 \;\;\forall v \in S   &\qquad &\text{\small($\ge 1$ executor per node)}
\end{alignedat}
\end{equation}
where $v$ is a node in logical plan $S$, $\mathcal{K}_v$ is its set of feasible configurations (determined by Configuration space in \S\ref{sec:overview-architecture}), $a_{v,k}$ is the GPU cost of node $v$ using configuration $k$, and $n_{v,k}$ is the number of executors with config $k$ assigned to node $v$.

Finally, the planner assigns per-type routing probabilities (Algorithm~\ref{algo:design-enumeration}, line~\ref{algline:routing}) at two levels: between executors in the same node and between alternative parallel paths in the logical plan.
At each level, routing probabilities for each request type sum to 1.
By default, routing probabilities are discretized with a step size of 0.1 (10\%), but finer granularities are supported at the cost of more physical plans to evaluate (\S\ref{sec:evaluation-sensitivity}).
The final result is a list of physical plans, each a complete specification for the runtime for deployment and execution.

\paragraph{Time complexity.}
Enumeration is exponential in the number of model components and colocatable edges, but since these are in practice at most tens (Qwen Omni has 6 components), enumeration time is manageable (\S\ref{sec:evaluation-overhead}).
The maximum number of subplans to merge ($k_c$) and to compose ($k_s$) bound the explosion---we vary these in Section~\ref{sec:evaluation-sensitivity}.

\subsection{Plan Evaluation and Selection}\label{sec:design-pruning}\label{sec:design-evaluation}

\begin{algorithm}[t]
  \algrule{}
  \KwIn{Physical plans $\mathcal{P}$\newline
    Executor throughput and latency profiles\newline
    Workload $\mathcal{W}$ with request type fractions $\{\pi_t\}$\newline
    Latency targets $\{L_t\}$ and headroom $\alpha$
  }
  \KwOut{Selected physical plan $d^*$}
  \algrule{}

  \Comment{Phase 1: Flow-based throughput estimation}
  Sort $\mathcal{P}$ by increasing executor capacity per node\;
  \For{\textnormal{each} $d \in \mathcal{P}$}{
    Find bottleneck node in $d$\label{algline:rate-matching-throughput}\;
    $R_d \leftarrow$ aggregate rate that saturates the bottleneck\;
    Drop plans with redundant capacity from $\mathcal{P}$\label{algline:prune-capacity}\;
  }

  \Comment{Phase 2: Latency estimation with Monte Carlo}
  \For{\textnormal{each} $d \in \mathcal{P}$}{
    Sample $K$ requests from $\mathcal{W}$\;
    \For{\textnormal{each request type} $t$}{
      Route type-$t$ requests through $d$\label{algline:rate-matching-latency}\;
      $F_{t,d} \leftarrow$ \textnormal{CDF of accumulated latencies}\;
      \Comment{$\alpha R_d \pi_t$ incoming rate, $F_{t,d}(L_t)$ met latency target}
      $G_{t,d} \leftarrow \alpha R_d \pi_t \cdot F_{t,d}(L_t)$\;
    }
  }
  Drop Pareto-suboptimal plans from $\mathcal{P}$\label{algline:prune}\;

  \Comment{Phase 3: Request-level simulation}
  \For{\textnormal{each} $d \in \mathcal{P}$}{
    $\{R_{t,d}, F_{t,d}\}_{t \in T} \leftarrow$ \textnormal{simulate} $d$ \textnormal{at rate} $\alpha \cdot R_d$\label{algline:simulate}\;
    $G_{t,d} \leftarrow R_{t,d} \cdot F_{t,d}(L_t)$ for each $t$\;
  }
  Drop Pareto-suboptimal plans from $\mathcal{P}$\;
  \Comment{Select: Default Policy is sum of goodput across types}
  $d^* \leftarrow \arg\max_d \mathrm{Policy}(G_{1,d}, \ldots, G_{|T|,d})$\label{algline:select}\;
  \Return{$d^*$}
  \algrule{}
  \caption{Three-phase evaluation and selection.}\label{algo:design-evaluation}
\end{algorithm}

Plan enumeration produces a large set of candidate physical plans.
The planner evaluates them in a coarse-to-fine pipeline (Algorithm~\ref{algo:design-evaluation}).
For all enumerated physical plans, a cheap network flow-based estimate provides accurate aggregate throughput estimates, and Monte Carlo sampling adds per-type latency estimates.
Pruning rules discard unpromising plans after each phase, and survivors are evaluated with a more expensive request-level simulator that refines the accuracy of per-type goodput estimates to select the final plan.
The pruning rules are exact; they only drop plans that are guaranteed to be redundant or suboptimal.

\paragraph{Network flow-based throughput estimation.}
Offline profiling provides the maximum throughput of each executor for each request type, and each node's throughput (i.e., capacity) is the sum of its executors' throughput.
Given the workload's per-type fractions $\pi_t$, each type contributes $R \cdot \pi_t$ to a node's load, split across executors according to routing probabilities.
Prior works have used (max) flow-based throughput analysis for a \emph{single} request type~\cite{distserve-osdi24,helix-asplos25}.
However, for A2A models, multiple request types flow through the graph simultaneously and share node capacity; at each node, the demands of all types passing through it are summed and compared against the node's capacity.
The node whose aggregate demand first reaches its capacity is the bottleneck, and the aggregate request rate $R_d$ at which this occurs is plan $d$'s maximum throughput (Algorithm~\ref{algo:design-evaluation}, line~\ref{algline:rate-matching-throughput}).
Plans that configure a node with more capacity than the node's aggregate demand are redundant, as they won't improve throughput or latency; all such plans are pruned (Algorithm~\ref{algo:design-evaluation}, line~\ref{algline:prune-capacity}).

$R_d$ is an ideal upper bound; system overheads and request bursts prevent it from being achieved in practice.
The planner therefore scales it down to $\alpha \cdot R_d$ (we find that the planner is robust to the setting of $\alpha$; \S\ref{sec:evaluation-sensitivity}), which is used as the incoming aggregate request rate for subsequent phases.

\paragraph{Latency estimation with Monte Carlo.}
The planner estimates request latency by Monte Carlo simulation (Algorithm~\ref{algo:design-evaluation}, line~\ref{algline:rate-matching-latency}): it randomly samples $K$ requests from the Workload, routes each through the plan, and accumulates per-executor processing latencies (sampled from profiling results) to produce per-type latency CDFs.
This is cheap because each request's latency is independent of others (i.e., queuing not modeled), but still provides a reasonable estimate because for well-balanced plans, queuing delay is bounded and latency is dominated by processing time.
After latency estimation, plans whose per-type goodput vectors are Pareto-dominated are pruned (Algorithm~\ref{algo:design-evaluation}, line~\ref{algline:prune}).

\paragraph{Request-level simulation.}
Earlier phases reasoned about aggregate flow and independent requests to provide good estimates for \emph{aggregate} goodput, but they are less precise for each request type because they do not precisely model queuing dynamics and inter-type interactions at shared nodes.
Candidates that survived earlier phases are therefore evaluated with a \emph{request-level simulator} (Algorithm~\ref{algo:design-evaluation}, line~\ref{algline:simulate}) that models the full request processing pipeline, refining accuracy to the individual request type-level.
The simulator runs the workload at rate $\alpha \cdot R_d$ through a physical plan and produces (1) a throughput estimate and (2) a latency CDF for each request type, from which goodput estimates per request type can be derived.
Plans that are Pareto-dominated on their goodput vector are pruned, and the final plan is selected (Algorithm~\ref{algo:design-evaluation}, line~\ref{algline:select}) according to a policy, which maximizes aggregate goodput by default.
Section~\ref{sec:evaluation-planner-accuracy} evaluates the accuracy of the pipeline.

\begin{figure}[t]
  \centering
  \includegraphics[width=0.35\textwidth]{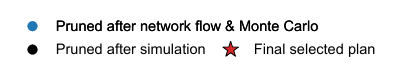}
  \vspace{-0.5em}

  \includegraphics[width=0.26\textwidth]{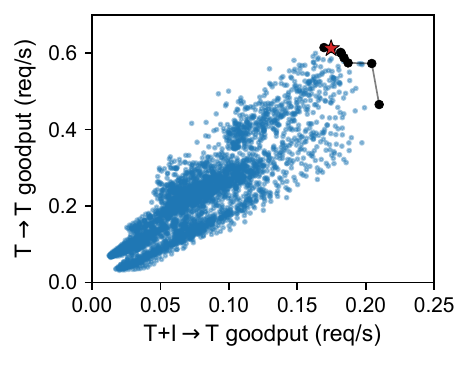}
  \vspace{-1.0em}
  \caption{
    Illustration of how physical plan evaluation, pruning, and selection work.
    Each dot represents a physical plan, and the two axes are goodput for two request types.
    Plans are progressively pruned after each phase (network flow, Monte Carlo, and simulation), reflected by the dots' color.
  }\label{fig:design-evaluation-pruning}
\end{figure}

Figure~\ref{fig:design-evaluation-pruning} visualizes the goodput of physical plans of a model with two request types, with colors showing at which phase the plan was pruned.
The plans pruned by network flow \emph{overlap} in goodput values with their non-redundant counterparts as they have the same bottleneck node, so it reduces the number of candidate plans significantly but not the visible point cloud.
Monte Carlo and request-level simulation eliminates Pareto-suboptimal plans.

\paragraph{Time complexity.}
Network flow and Monte Carlo are both linear in the number of candidate physical plans, and cheap per plan; the expensive request-level simulation runs only on the small set of survivors after pruning.
Section~\ref{sec:evaluation-overhead} breaks down the number of plans and time spent in each phase.

\subsection{Adapting to Changes}\label{sec:design-adapting}

Three types of changes can occur in a deployment scenario, each requiring progressively more work to adapt.

\paragraph{GPU budget changes.}
If the GPU budget $N$ changes but the workload distribution and model remain the same, the planner can simply re-run with the new $N$.
Profiling results remain valid, and the plan enumeration and evaluation phases are fast enough to re-run semi-online (\S\ref{sec:evaluation-overhead}).

\paragraph{Workload changes.}
If request types fractions $\pi_t$ change, profiling results can be reused because the profiler records per-request latency measurements.
Changing the request type distribution is a matter of re-weighting existing samples.
The planner re-runs enumeration and evaluation with the updated workload fractions.
On the other hand, if new request types are added, or when the characteristics of existing request types change (e.g., significantly different sequence lengths), the new/updated type needs to be re-profiled.
We show this in Section~\ref{sec:evaluation-workload-drift}.

\paragraph{Model or hardware changes.}
If the model architecture changes (e.g., a new component is added) or the hardware changes (e.g., migrating to a different GPU type), profiling must be re-run from scratch, as per-component throughput and latency characteristics are no longer valid.
This is the most expensive adaptation, but it would also be the least frequent in practice.

  \section{Implementation}\label{sec:implementation}

Cornfigurator is designed to be runtime-agnostic; it works with runtimes ranging from those supporting only a single executor type to fully generic A2A serving runtimes.
As a proof of concept, we implemented Cornfigurator on top of Cornserve~\cite{cornserve-github, cornserve-cais26}, a distributed A2A model serving platform.

\subsection{Profiler}\label{sec:impl-profiler}

The profiler benchmarks each supported executor under the workload to collect the throughput and latency data that the planner uses.
The profiler deploys executors on the target hardware via the runtime and sends requests, sweeping over configuration knobs (e.g., batch sizes, parallelism degrees).
Profiling is performed at a saturating request rate and measured only during the \emph{steady state} window of the engine, excluding ramp up and down periods.
In order to extract pure processing time, queuing delay is subtracted from end-to-end latency measurements using runtime traces.

\subsection{Planner}\label{sec:impl-planner}

The planner is implemented in about 5K lines of Rust.
The plan evaluation phases (\S\ref{sec:design-pruning}) implement the following extra optimizations to improve accuracy.
The accuracy of Cornfigurator's planner is evaluated in Section~\ref{sec:evaluation-planner-accuracy}.

\paragraph{Occupancy-aware latency scaling.}
Flow-based estimation finds the bottleneck node in the plan that bounds the whole plan's throughput.
All other nodes are non-bottlenecks, and likely operate at a batch size smaller than its \emph{configured} batch size, which lead to lower latency than what the profiler measured at the configured batch size.
Therefore, during Monte Carlo simulation, the planner scales latency numbers for non-bottleneck nodes to reflect their effective batch size.

\paragraph{Modeling CPU-GPU overlap.}
Multimodal encoders perform CPU-GPU pipelining: the CPU preprocesses input data (e.g., image decoding) while the GPU runs the encoder model.
When a request arrives at an idle multimodal encoder executor, it experiences the full CPU + GPU time.
However, when the encoder is already busy, CPU preprocessing happens while the request is in the queue, and processing time reduces to GPU time alone.
This difference is significant especially when batch size is small.
Thus, the simulator checks whether the executor is idle or busy and applies the corresponding processing time (CPU + GPU, or GPU alone).

\looseness=-1
\paragraph{Accounting for communication overhead.}
Intermediate tensors must be transferred between disaggregated executors.
In our setup, the median transfer latency across representative tensor sizes and traffic rates is approximately 10 ms.
While this transfer latency is not the bottleneck as computation dominates, the planner can account for it by adding this median delay at each disaggregated edge in the plan.

	\section{Evaluation}\label{sec:evaluation}

We plan the deployment of recent A2A models with diverse architectures, modalities, and request types with Cornfigurator and compare with state-of-the-art baselines.
We find:
\begin{denseitemize}
  \item Cornfigurator produces high-quality deployment plans for generic A2A models that achieve 1.24$\times$--2.73$\times$ higher goodput than baselines (\S\ref{sec:evaluation-omni}).
  \item Cornfigurator produces plans that match or deliver 1.12$\times$--6.32$\times$ higher goodput over baselines for special case A2A models: multimodal input (\S\ref{sec:evaluation-mllm}) and output (\S\ref{sec:evaluation-qwen-image}) models.
  \item Cornfigurator's planner estimates goodput with high accuracy (\S\ref{sec:evaluation-planner-accuracy}), supports adapting to workload drift (\S\ref{sec:evaluation-workload-drift}), is robust to parameter choices (\S\ref{sec:evaluation-sensitivity}), and produces plans with reasonable overhead (\S\ref{sec:evaluation-overhead}).
\end{denseitemize}

\subsection{Common Evaluation Setup}
\label{sec:evaluation-setup}

Different A2A models have different input and output modalities, and thus different representative workloads and baselines.
We describe common setup across all evaluations here, and architecture-specific setup in respective subsections.

\paragraph{Testbed.}
All experiments are run on two AWS p4de.24xlarge instances, each with $8\times$ NVIDIA A100-80GB GPUs connected via NVSwitch and 400 Gbps cross-node bandwidth.

\paragraph{Baselines.}
Baselines include plans generated by experts/ developers of vLLM~\cite{pagedattention-sosp23}, vLLM-Omni~\cite{vllm-omni-github}, ModServe~\cite{modserve-socc26}, and EPD~\cite{epd-icml25}.\footnote{SGLang-Omni~\cite{sglang-omni-github} is not included as it is in early development stage, falling short on stability, model support, and deployment plans at the moment.}
Cornfigurator is a strict superset of all baselines, so the plans that come with each baseline system can be expressed as a Cornfigurator plan.
We deploy and evaluate all plans on Cornserve~\cite{cornserve-github, cornserve-cais26}, a generic A2A serving platform that natively supports Cornfigurator's plan space to remove performance differences due to implementation differences~\cite{llmperfeval-arxiv25}.

\paragraph{Metric.}
Our primary performance metric is goodput (\S\ref{sec:overview-objective}): the throughput of requests that meet their per-type latency targets.
We compare the \emph{maximum achievable} goodput of each plan, which is the goodput achieved by the highest request rate that the plan can support.

\paragraph{Latency targets.}
We set the latency target of each request type (\S\ref{sec:overview-objective}) in a manner that reflects the raw amount of computation requirements of that type, while still being achievable by a shared deployment plan that serves all types simultaneously.
Unless mentioned otherwise, we identify the highest-throughput configuration for each request type served in isolation from offline profiling results and take its p25 latency as the target for that type.
We show the sensitivity of goodput to the choice of percentile in Section~\ref{sec:evaluation-sensitivity}.

\subsection{Multimodal Input, Multimodal Output}
\label{sec:evaluation-omni}

\begin{figure}[t]
  \centering
  \includegraphics[width=0.85\columnwidth]{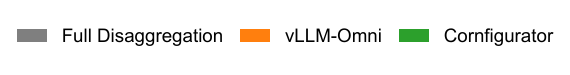}
  \vspace{-2em}

  \subfloat[1/3 audio output]{
    \includegraphics[width=0.21\textwidth]{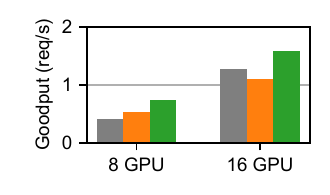}
    \label{fig:evaluation-omni-goodput-less-audio}
  }%
  \subfloat[2/3 audio output]{
    \includegraphics[width=0.21\textwidth]{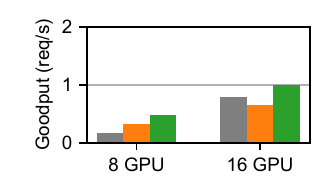}
    \label{fig:evaluation-omni-goodput-more-audio}
  }
  \vspace{-0.5em}
  \caption{
    Maximum achievable goodput of Qwen 3 Omni under the ServeGen workload on 8 and 16 GPUs with (a) 1/3 and (b) 2/3 audio output requests.
  }
  \label{fig:evaluation-omni-goodput}
\end{figure}

\paragraph{Setup.}
We evaluate Qwen 3 Omni 30B~\cite{qwen3-omni-arxiv25} with the ServeGen~\cite{servegen-nsdi26} workload, which includes request timestamps.
The workload has 8 request types: \{T+I, T+I+V, T+I+A, T+I+V+A\} $\to$ \{T, A\}, where inputs always include text and image, and include video and/or audio input half of the time.
We evaluate two audio output fractions (1/3 and 2/3) to study how the workload's modality mix affects plans and goodput.

\paragraph{Baselines.}
We compare against two baselines.
First is vLLM-Omni~\cite{vllm-omni-github}, which uses an expert-tuned but fixed deployment strategy.
Second is Full Disaggregation, which is a restricted version of Cornfigurator; its graph-level configuration space is fixed to fully disaggregated (i.e., each model component running in a separate executor), but otherwise follows the full Cornfigurator planning pipeline to optimize executor-level configurations and GPU allocation.

\paragraph{Goodput comparison.}
Figure~\ref{fig:evaluation-omni-goodput} compares goodput across systems on 8 and 16 GPUs under both audio output fractions.
Cornfigurator achieves 1.24$\times$--1.84$\times$ higher goodput than baselines across both GPU budgets with 1/3 audio output (Figure~\ref{fig:evaluation-omni-goodput-less-audio}), and the gains widen to
1.25$\times$--2.73$\times$ with 2/3 audio (Figure~\ref{fig:evaluation-omni-goodput-more-audio}).
As the audio generator becomes a heavier bottleneck, Cornfigurator's ability to allocate resources proportionally to component demand becomes more valuable.

Scaling from to 16 GPUs, both Cornfigurator and Full Disaggregation achieve super-linear scaling because the larger budget allows better component balance.
However, Full Disaggregation still falls short of Cornfigurator because it disaggregates all components including those with less load, leading to underutilized GPU resource.
In contrast, Cornfigurator automatically finds efficient colocation strategies.

\begin{figure}[t]
  \centering
  \includegraphics[width=0.9\columnwidth]{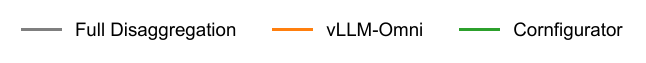}
  \vspace{-1.5em}

  \subfloat[Text output]{
    \includegraphics[width=0.21\textwidth]{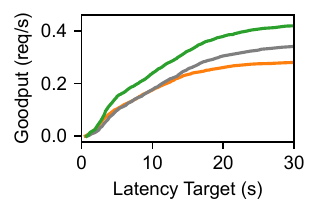}
    \label{fig:evaluation-omni-slo-tiva-t}
  }%
  \subfloat[Audio output]{
    \includegraphics[width=0.21\textwidth]{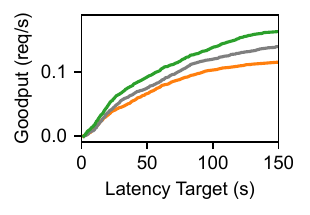}
    \label{fig:evaluation-omni-slo-tiva-a}
  }
  \vspace{-0.6em}
  \caption{Goodput of requests in Qwen 3 Omni with T+I+V+A input with (a) text output and (b) audio output vs. latency targets under 1/3 audio workload on 16 GPUs.}
  \label{fig:evaluation-omni-slo-pertype}
\end{figure}

\paragraph{Per-type goodput vs.\ latency target.}
Figure~\ref{fig:evaluation-omni-slo-pertype} shows the goodput of two request types in Cornfigurator with 1/3 audio, 16 GPU deployment (remaining six types are in Appendix~\ref{sec:appendix-omni-slo}).
Cornfigurator's plans achieve higher goodput than baselines across the full range of latency targets, from tight to relaxed.

\begin{figure}[t]
  \centering
  \subfloat[Model definition]{%
    \includegraphics*[width=0.30\textwidth]{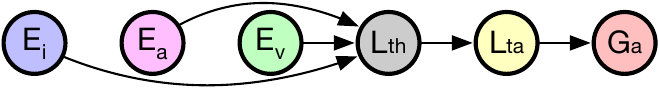}%
    \label{fig:evaluation-omni-plan-model}%
  }
  \vspace{-0.3em}

  \subfloat[Physical plan]{%
    \includegraphics[width=0.35\textwidth]{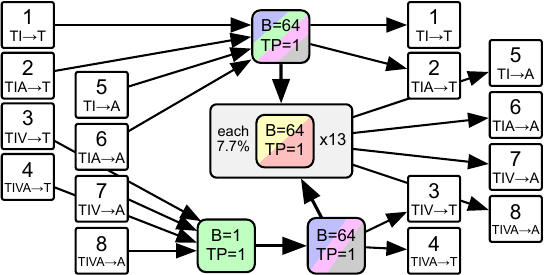}%
    \label{fig:evaluation-omni-plan-physical}%
  }
  \vspace{-0.3em}
  \caption{
    (a) Qwen 3 Omni definition graph and (b) Cornfigurator's deployment plan under 2/3 audio workload on 16 GPUs.
    The planner prepared a video-only subplan (with a disaggregated video encoder) for heavy video-input requests.
  }
  \label{fig:evaluation-omni-plan}
\end{figure}

\paragraph{Physical plans.}
We examine the plans generated by Cornfigurator to understand its planning decisions.
On the 1/3 audio workload (executor configs are similar and thus omitted):
\begin{denseitemize}
\item \textbf{8 GPUs}: $1\times$ (E$_\text{img}$) + $2\times$ (E$_\text{vid}$+E$_\text{aud}$+L$_\text{th}$) + $5\times$ (L$_\text{ta}$+G$_\text{aud}$)
\item \textbf{16 GPUs}: $1\times$ (E$_\text{aud}$) + $4\times$ (E$_\text{img}$+E$_\text{vid}$+L$_\text{th}$) + $11\times$ (L$_\text{ta}$+G$_\text{aud}$)
\end{denseitemize}
Cornfigurator switches from disaggregating the image encoder on 8 GPUs to disaggregating the audio encoder on 16 GPUs, as the additional GPU budget allowed it to realize better balance.
On the 2/3 audio workload, the 8 GPU plan is $1\times$ (E$_\text{img}$+E$_\text{vid}$+E$_\text{aud}$+L$_\text{th}$) + $6\times$ (L$_\text{ta}$) + $1\times$ (G$_\text{aud}$), reflecting that increased audio output load promoted talker LLM and vocoder disaggregation.
With 16 GPUs (Figure~\ref{fig:evaluation-omni-plan}), Cornfigurator has more GPUs to express higher degrees of specialization: the plan originates from a compound logical subplan (\S\ref{sec:design-enumeration}) that merges two simple subplans sharing $13\times$ (L$_\text{ta}$+G$_\text{aud}$).
One subplan serves all heavy video-input requests with a disaggregated video encoder, while the other subplan serves the rest of the requests with a monolithic configuration.

\subsection{Multimodal Input, Text Output}
\label{sec:evaluation-mllm}

\paragraph{Setup.}
MLLMs are a special case of A2A models with multimodal input and text-only output.
We evaluate on two MLLMs, InternVL 3 38B~\cite{internvl3-arxiv25} and Qwen 3 VL 32B~\cite{qwen3-vl-arxiv25}, under the ServeGen~\cite{servegen-nsdi26} workload, which includes request timestamps.
Half of the requests have image input by default, and we vary this portion in controlled experiments (\S\ref{sec:evaluation-workload-drift}).

\paragraph{Baselines.}
We compare Cornfigurator plans with vLLM~\cite{pagedattention-sosp23} and EPD~\cite{epd-icml25}, which are restricted versions of Cornfigurator with fixed graph-level configurations (monolithic for vLLM, encoder--prefill--decode disaggregation for EPD) and follow the full Cornfigurator planning pipeline for executor-level configurations and GPU allocation, and ModServe~\cite{modserve-socc26} (TP=4, equal encoder/LLM GPU split).\footnote{vLLM-Omni is built for multimodal \emph{output} models, so it is not included.}

\begin{figure}[t]
  \centering
  \includegraphics[width=0.85\columnwidth]{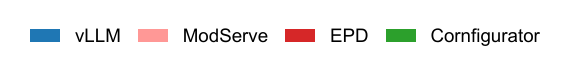}
  \vspace{-2em}

  \subfloat[InternVL 3 38B]{
    \includegraphics[width=0.21\textwidth]{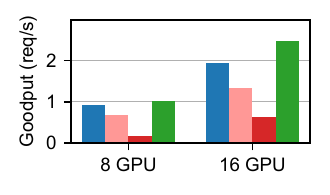}
    \label{fig:evaluation-mllm-goodput-intern}
  }%
  \subfloat[Qwen 3 VL 32B]{
    \includegraphics[width=0.21\textwidth]{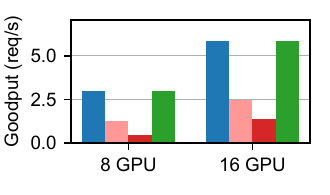}
    \label{fig:evaluation-mllm-goodput-qwen}
  }
  \vspace{-0.6em}
  \caption{Maximum achievable goodput of two MLLMs under the ServeGen workload on 8 and 16 GPUs.}
  \label{fig:evaluation-mllm-goodput}
\end{figure}

\begin{figure}[t]
  \centering
  \vspace{0.7em}
  \includegraphics[width=0.35\textwidth]{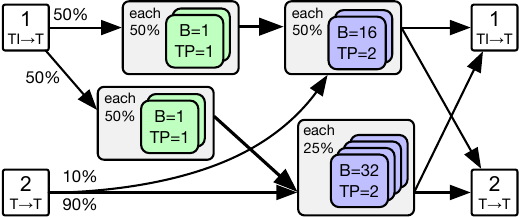}
  \caption{
    Cornfigurator's deployment plan for InternVL 3 38B (\raisebox{-0.5ex}{\includegraphics[height=1.0em]{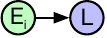}}) under 50\% image workload on 16 GPUs.
  }
  \label{fig:evaluation-multimodal-input-servegen}
\end{figure}

\paragraph{Goodput comparison.}
Figure~\ref{fig:evaluation-mllm-goodput} compares goodput on 8 and 16 GPUs.
For InternVL 3 (Figure~\ref{fig:evaluation-mllm-goodput-intern}), Cornfigurator achieves 1.12$\times$--6.06$\times$ higher goodput compared to baselines on 8 GPUs and 1.27$\times$--3.95$\times$ on 16 GPUs.
Figure~\ref{fig:evaluation-multimodal-input-servegen} illustrates the 16 GPU physical deployment plan: the planner disaggregates the large 6B image encoder onto dedicated GPUs.
For Qwen 3 VL (Figure~\ref{fig:evaluation-mllm-goodput-qwen}), Cornfigurator matches or exceeds all baselines (1.00$\times$--6.32$\times$), correctly identifying that disaggregation does not help for this model's small encoder.
EPD performs poorly because its advantages generally show with short output lengths~\cite{epd-icml25}.

\begin{figure}[t]
  \centering
  \includegraphics[width=0.9\columnwidth]{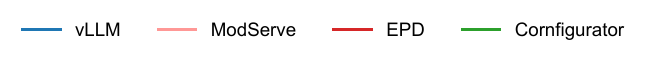}
  \vspace{-1.5em}

  \subfloat[Text-only input]{
    \includegraphics[width=0.21\textwidth]{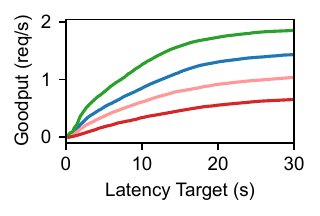}
    \label{fig:evaluation-mllm-slo-t-t}
  }%
  \subfloat[Image+text input]{
    \includegraphics[width=0.21\textwidth]{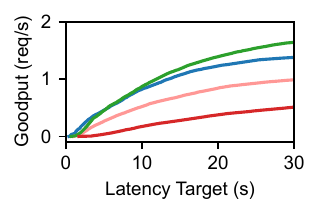}
    \label{fig:evaluation-mllm-slo-ti-t}
  }
  \vspace{-0.6em}
  \caption{Goodput of requests in InternVL 3 with (a) text-only input and (b) image+text input vs.\ latency targets under 50\% image workload on 16 GPUs.}
  \label{fig:evaluation-mllm-slo-pertype}
\end{figure}

\paragraph{Per-type goodput vs.\ latency target.}
Figure~\ref{fig:evaluation-mllm-slo-pertype} shows that Cornfigurator achieves higher goodput than baselines for both InternVL 3 request types across most latency targets.

\subsection{Text Input, Multimodal Output}
\label{sec:evaluation-qwen-image}

\paragraph{Setup and baseline.}
We evaluate Qwen-Image~\cite{qwen-image-arxiv25} using DiffusionDB~\cite{diffusionDB-acl23} prompts with a mixture of low (512$\times$512) and high (1024$\times$1024) resolution requests\JW{with a ratio of @Jeff}~\cite{tetriserve-asplos26}, submitted with Poisson intervals.
Cornfigurator's plan is compared against Full Disaggregation and vLLM-Omni~\cite{vllm-omni-github}'s plan.

\begin{figure}[t]
  \centering
  \includegraphics[width=0.85\columnwidth]{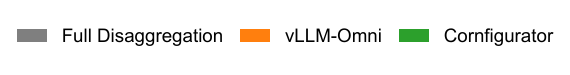}
  \includegraphics[width=0.32\textwidth]{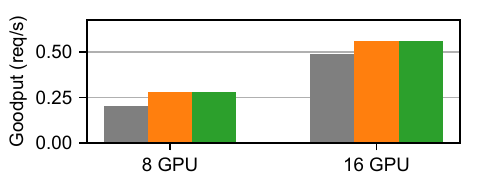}
  \vspace{-1.0em}
  \caption{
    Maximum achievable goodput of Qwen-Image under the DiffusionDB workload on 8 and 16 GPUs.
  }\label{fig:evaluation-qwen-image}
\end{figure}

\paragraph{Goodput comparison.}
Figure~\ref{fig:evaluation-qwen-image} compares goodput of Qwen-Image on 8 and 16 GPUs.
Qwen-Image is a simple two-component model, and the first component (LLM prefill) is computationally lightweight compared to the second component (DiT).
As such, disaggregating the two components does not yield benefits as the first component underutilizes its dedicated GPUs, so Cornfigurator correctly identifies and uses a monolithic deployment plan.
Appendix~\ref{sec:appendix-qwen-image-slo} shows the goodput vs.\ latency target curve for this model.

\subsection{Planner Accuracy}\label{sec:evaluation-planner-accuracy}

\begin{table}[t]
  \footnotesize
  \vspace{1.0em}
  \centering
  \begin{tabular}{llrrrr}
    \toprule
    \multirow{2}{*}{\textbf{Input}} & \multirow{2}{*}{\textbf{Output}} & \multicolumn{2}{c}{\textbf{Mean Abs.\ Err. (\%)}} & \multicolumn{2}{c}{\textbf{Median Abs.\ Err. (\%)}} \\
    \cmidrule(lr){3-4} \cmidrule(lr){5-6}
                                    & & F\&M & Simulator & F\&M & Simulator \\
    \midrule
    T+I     & T & $13.0$ & $5.8$ & $8.5$ & $4.9$ \\
    T+I+V   & T & $17.5$ & $5.4$ & $17.2$ & $4.0$ \\
    T+I+A   & T & $12.9$ & $6.5$ & $11.5$ & $4.4$ \\
    T+I+V+A & T & $16.9$ & $9.3$ & $11.6$ & $7.2$ \\
    T+I     & A & $9.6$ & $5.7$ & $3.1$ & $4.6$ \\
    T+I+V   & A & $12.0$ & $5.7$ & $5.9$ & $3.2$ \\
    T+I+A   & A & $15.6$ & $5.1$ & $14.5$ & $3.4$ \\
    T+I+V+A & A & $12.2$ & $4.9$ & $6.8$ & $4.0$ \\
    \midrule
    \multicolumn{2}{l}{Aggregated} & $10.7$ & $4.1$ & $8.6$ & $3.3$ \\
    \bottomrule
  \end{tabular}
  \vspace{0.7em}
  \caption{Per-type and aggregate goodput prediction error (\%) for Qwen 3 Omni on 8 and 16 GPUs. F\&M is flow-based throughput estimation \& Monte Carlo latency estimation phases combined, and Simulator is request-level simulation.}
  \label{tab:evaluation-planner-accuracy}
  \vspace{-1em}
\end{table}

We evaluate the accuracy of the goodput predictions of the plan evaluation pipeline (\S\ref{sec:design-evaluation}), namely the cheap flow-based throughput estimation + Monte Carlo latency estimation (\emph{F\&M}) and the more expensive request-level simulator (\emph{Simulator}), by comparing predictions with real serving measurements.
Table~\ref{tab:evaluation-planner-accuracy} reports prediction accuracies across Qwen 3 Omni experiments on 8 and 16 GPUs.
\emph{F\&M} is sufficient for pruning based on aggregate goodput because it is accurate when it comes to aggregate goodput across all request types (mean absolute error 10.7\%, median 8.6\%).
However, it has overall higher error for per-request-type goodput as it does not model queuing effects and inter-type contention at shared executors.
On the other hand, \emph{Simulator} is not only more accurate in aggregate goodput (mean absolute error 4.1\%, median 3.3\%), but also significantly more accurate for individual request types, making it essential for refining the evaluation of plans that are closer to the final plan.

\begin{table}[t]
  \footnotesize
  \centering
  \begin{tabular}{lrrrr}
    \toprule
    \textbf{Image prob.} & \textbf{vLLM} & \textbf{ModServe} & \textbf{EPD} & \textbf{Cornfigurator} \\
    \midrule
    25\% & 2.49 & 1.92 & 0.89 & 3.60 \\
    50\% & 1.95 & 1.33 & 0.63 & 2.48 \\
    75\% & 1.75 & 1.71 & 0.35 & 1.75 \\
    \bottomrule
  \end{tabular}
  \vspace{0.5em}
  \caption{InternVL 3 goodput (req/s) under ServeGen with varying image probability on 16 GPUs.}
  \label{tab:evaluation-mllm-servegen-imgprob}
  \vspace{-1.0em}
\end{table}

\subsection{Workload Drift}\label{sec:evaluation-workload-drift}

Cornfigurator helps runtimes to adapt to workload drift (\S\ref{sec:design-adapting}).
To demonstrate this, we vary the image input probability of the ServeGen workload for InternVL 3 from 25\% to 75\% on 16 GPUs, and let Cornfigurator re-plan for each workload.
Table~\ref{tab:evaluation-mllm-servegen-imgprob} shows that Cornfigurator consistently matches or outperforms baselines across all image probabilities by adapting to the changing workload.
As expected, the planner does not require any re-profiling to adapt to ratio changes, so re-planning takes single- to at most double-digit seconds.
Appendix~\ref{sec:appendix-workload-drift} provides visualizations of Cornfigurator's plans at different image probabilities.

\subsection{Sensitivity Analysis}\label{sec:evaluation-sensitivity}

\begin{table}[t]
  \footnotesize
  \centering
  \begin{tabular}{rrr}
    \toprule
    $\alpha$ & \textbf{Goodput (req/s)} & \textbf{Prediction error} \\
    \midrule
    0.5 & 1.08 & $-0.6$\% \\
    0.6 & 1.59 & $-1.8$\% \\
    0.7 & 1.83 & $+0.8$\% \\
    0.8 & 1.89 & $-1.6$\% \\
    \bottomrule
  \end{tabular}
  \vspace{0.5em}
  \caption{Effect of throughput headroom $\alpha$ on goodput and planner accuracy. $\alpha$ scales down the rate-matched maximum throughput $R_d$ to $\alpha \cdot R_d$ before latency estimation (\S\ref{sec:design}).}
  \label{tab:evaluation-sensitivity-alpha}
  \vspace{-1em}
\end{table}

Cornfigurator's planner has several parameters as part of its design.
In this section, we vary these parameters to understand their effect on goodput and the planning process.

\paragraph{Throughput headroom.}
After flow-based throughput estimation, the planner scales down the ideal maximum throughput $R_d$ by a headroom factor $\alpha$ (\S\ref{sec:design-evaluation}).
Table~\ref{tab:evaluation-sensitivity-alpha} shows that the planner's prediction error remains within $\pm 2\%$ across all tested values of $\alpha$, demonstrating robustness.

\begin{figure}[t]
  \vspace{-0.8em}
  \centering
  \subfloat[Latency target percentile]{
    \includegraphics[width=0.21\textwidth]{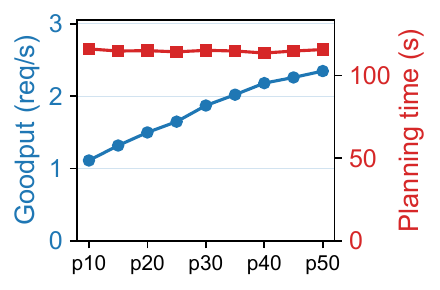}
    \label{fig:evaluation-omni-goodput-slo}
  }
  \subfloat[Routing step]{
    \includegraphics[width=0.21\textwidth]{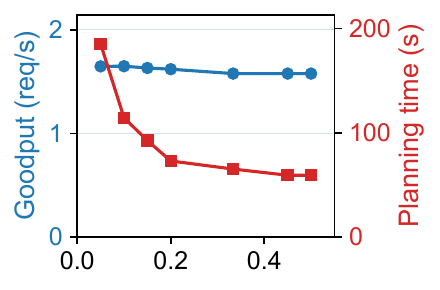}
    \label{fig:evaluation-sensitivity-delta}
  }

  \vspace{-0.5em}
  \subfloat[Merge limit $k_c$]{
    \includegraphics[width=0.21\textwidth]{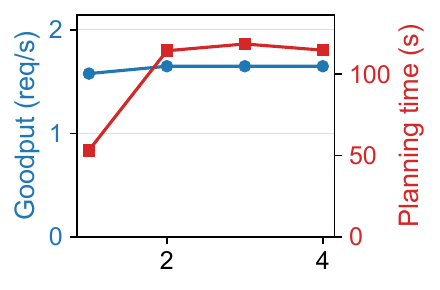}
    \label{fig:evaluation-sensitivity-kc}
  }
  \subfloat[Composition limit $k_s$]{
    \includegraphics[width=0.21\textwidth]{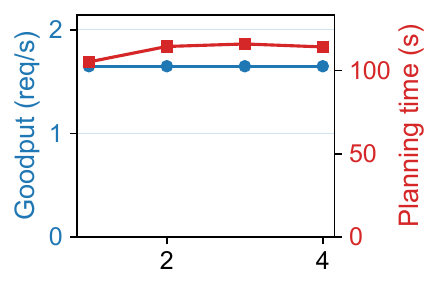}
    \label{fig:evaluation-sensitivity-ks}
  }
  \vspace{-0.5em}
  \caption{
    Sensitivity analysis with Qwen 3 Omni on 16 GPUs.
    Each plot shows planner goodput (left y-axis) and planning time (right y-axis) as one parameter is varied.
  }\label{fig:evaluation-sensitivity}
  \vspace{0.5em}
\end{figure}

\paragraph{Latency target percentile.}
Figure~\ref{fig:evaluation-omni-goodput-slo} shows goodput as the latency target percentile is swept from p10 (tight) to p50 (relaxed).
As the target loosens, the planner can find higher throughput plans with weaker latency constraints, increasing goodput without increasing planning time.
Our default p25 (\S\ref{sec:evaluation-setup}) represents a practical middle ground.

\paragraph{Routing probability discretization.}
When annotating the logical plan with routing probabilities, the planner discretizes probabilities into fixed steps (\S\ref{sec:design-evaluation}).
A finer step size would unlock higher goodput plans but also increases planning time, as shown by Figure~\ref{fig:evaluation-sensitivity-delta}.
We observe that a step size of 0.1 (10\% increments) is sufficient to capture goodput benefits without increasing planning time excessively.

\paragraph{Subplan merge and composition limits.}
During logical plan enumeration, up to $k_c$ simple subplans are merged into compound subplans and up to $k_s$ subplans into a logical plan (\S\ref{sec:design-enumeration}).
Figures~\ref{fig:evaluation-sensitivity-kc} and~\ref{fig:evaluation-sensitivity-ks} confirm that our default $k_c = k_s = 2$ captures most of the benefits of composition with manageable planning overhead.

\subsection{Cornfigurator Overhead}\label{sec:evaluation-overhead}

Profiling and planning are one-time costs per (model, workload) pair, amortized over the deployment lifetime.

\paragraph{Profiling.}
Profiling averages 4.2 hours on a single node per (model, workload) pair across our evaluation suite.
Qwen-Image is the cheapest to profile (27 configurations, 2.6 node-hours) due to its simpler two-component architecture, while Qwen 3 Omni is the most expensive (128 configurations, 9.9 node-hours) due to the number of components and types.

\begin{table}[t]
  \footnotesize
  \centering
  \begin{tabular}{lrrr}
    \toprule
    \textbf{Phase} & \textbf{Time (s)} & \textbf{Survivors} & \textbf{Fraction} \\
    \midrule
    Plan enumeration                      & 77.25  & 483.41M & 100\% \\
    Network flow for throughput           & 3.48   & 1.95M   & 0.40\% \\
    Monte Carlo for latency               & 34.23  & 25      & $<$0.01\% \\
    Request-level simulation              & 0.83   & 5       & $<$0.01\% \\
    \bottomrule
  \end{tabular}
  \vspace{0.5em}
  \caption{Planning time breakdown (Qwen 3 Omni, 16 GPUs). Each phase prunes candidates before moving on to the next.}
  \label{tab:evaluation-planner-speedup}
  \vspace{-1em}
\end{table}

\paragraph{Planning.}
Table~\ref{tab:evaluation-planner-speedup} shows how long each phase of the planning pipeline takes for Qwen 3 Omni on 16 GPUs, and how many candidates survive after each phase.
This is the most complex planning case in our evaluation suite, and the planner produces a final set of 5 plans from an initial candidate space of nearly 500 million.
Without pruning, running request-level simulation on all candidates would take over 4,400 hours; the coarse-to-fine pipeline reduces this to under 2 minutes of total planning time, with only 25 plans (${<}0.01\%$) reaching simulation.

	\section{Related Work}
\label{sec:related}

\paragraph{Serving generic and special cases of A2A models.}
vLLM-Omni~\cite{vllm-omni-github} and SGLang-Omni~\cite{sglang-omni-github} are serving runtimes that support generic A2A models, but they do not provide an automated planner.
Cornfigurator fills this gap by automatically searching over the space of colocation, disaggregation, and configuration decisions.
On the other hand, a large body of work targets special cases of A2A models.
For LLM and MLLM serving, many provide serving engines, disaggregation mechanisms, and/or deployment planners~\cite{pagedattention-sosp23,sglang-arxiv25,distserve-osdi24,splitwise-isca24,sarathiserve-osdi24,andes-arxiv24,helix-asplos25,modserve-socc26,epd-icml25,nvidia-dynamo,aiconfigurator-arxiv26,nanoflow-osdi25,prefillonly-sosp25,jenga-sosp25}, but do not generalize to generic A2A models.
For multimodal generation, existing works on serving optimize inference within a single engine for a single component~\cite{xdit-arxiv24,nirvana-nsdi24,tetriserve-asplos26}.
Each of these promote a specific deployment strategy for its target model family.
Cornfigurator instead treats these strategies as points in a search space and selects the best combination for the given model and workload.

\paragraph{Resource allocation and parallelization for training.}
Cornfigurator's deployment planning problem is analogous to large model parallelization for training.
Early systems~\cite{gpipe-neurips19,megatronlm-sc21,deepspeed-billionparam} relied on manual, expert-designed strategies, akin to monolithic or fixed disaggregated strategies in serving.
Automated planners emerged to search over parallelization strategies~\cite{alpa-osdi22,gspmd-arxiv21,oobleck-sosp23,wlb-llm-osdi25,sailor-sosp25,dcp-sosp25,cornstarch-arxiv25}, analogous to Cornfigurator's planner that automatically explores the space of deployment strategies rather than unilaterally prescribing one.

	\section{Conclusion}\label{sec:conclusion}

We present Cornfigurator, the first automated deployment planner for generic Any-to-Any model inference serving.
Cornfigurator surfaces the characteristics and requirements of each request type, instead of lumping them together into a single, aggregate objective and constraint.
To do so, Cornfigurator systematically explores the vast space of deployment configurations and uses coarse-to-fine statistical evaluation to select the final plan.
As more models aim to align with the multimodal nature of human perception and interaction, we believe Cornfigurator's automated planner will facilitate efficient model development and deployment at scale.

  \label{EndOfPaper}

  \bibliographystyle{plain}
  \bibliography{ref}
	\clearpage

	\appendix

\section{Compute-Proportional Deadlines}\label{sec:appendix-normalization}

For each request type $t$, let $\ell_t$ denote the sum of per-component compute latencies along the computation path of type $t$.
This quantity is determined by the model architecture and hardware, independent of the deployment plan.
Any request's latency can be written as $\ell_r = \ell_t \cdot X_t$, where $X_t = \ell_r / \ell_t$ captures queuing delay, load-dependent effects, and distributional shape.
Per-type deadlines are set as $L_t = L \cdot \ell_t / \ell_{\max}$, where $\ell_{\max} = \max_t \ell_t$.

\begin{theorem}\label{thm:normalization}
Under per-type deadlines $L_t = L \cdot \ell_t / \ell_{\max}$, the per-type CDF at the deadline threshold is
\begin{equation}
  \Pr(\ell_r \le L_t \mid t_r = t) = F_{X_t}(L / \ell_{\max}),
\end{equation}
where $F_{X_t}$ is the CDF of $X_t$.
That is, all types evaluate their shape CDF at the same argument $L / \ell_{\max}$, independent of their scale $\ell_t$.
\end{theorem}

\begin{proof}
Substituting $\ell_r = \ell_t \cdot X_t$ and $L_t = L \cdot \ell_t / \ell_{\max}$:
\begin{align}
  \Pr(\ell_r \le L_t \mid t_r = t)
  &= \Pr\!\left(\ell_t \cdot X_t \le L \cdot \frac{\ell_t}{\ell_{\max}}\right) \nonumber\\
  &= \Pr(X_t \le L / \ell_{\max}) \nonumber\\
  &= F_{X_t}(L / \ell_{\max}).
\end{align}
The scale $\ell_t$ cancels, leaving only the shape distribution $X_t$.
\end{proof}

Without per-type deadlines (i.e., a single global threshold $L$ for all types), the per-type CDF is $F_{X_t}(L / \ell_t)$, where the argument $L / \ell_t$ varies across types by an order of magnitude (e.g., $L/4.5$ vs.\ $L/18$ for text and audio output).
Per-type deadlines eliminate this scale dependence entirely.

\begin{corollary}\label{cor:identical-cdf}
If all request types share the same distributional shape ($X_t \stackrel{d}{=} X$ for all $t$), then all per-type CDFs evaluated at their respective deadlines are identical, and the percentile constraint binds equally on every request type.
\end{corollary}

\section{Qwen 3 Omni Per-Type Goodput vs.\ Latency Target}\label{sec:appendix-omni-slo}

\begin{figure}[p]
  \centering
  \includegraphics[width=0.85\columnwidth]{figures/goodput_slo_16_gpu_omni_ra0.3_latencies/legend.pdf}
  \vspace{-0.5em}

  \subfloat[TI$\to$T]{
    \includegraphics[width=0.46\columnwidth]{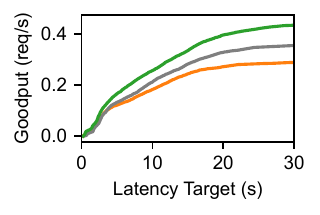}
  }%
  \subfloat[TIV$\to$T]{
    \includegraphics[width=0.46\columnwidth]{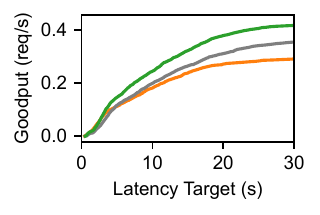}
  }

  \vspace{-0.5em}
  \subfloat[TIA$\to$T]{
    \includegraphics[width=0.46\columnwidth]{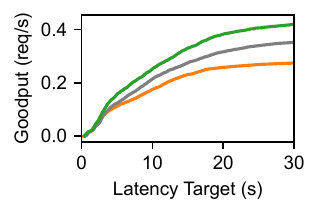}
  }%
  \subfloat[TIVA$\to$T]{
    \includegraphics[width=0.46\columnwidth]{figures/goodput_slo_16_gpu_omni_ra0.3_latencies/TIVA_T.pdf}
  }

  \vspace{-0.5em}
  \subfloat[TI$\to$A]{
    \includegraphics[width=0.46\columnwidth]{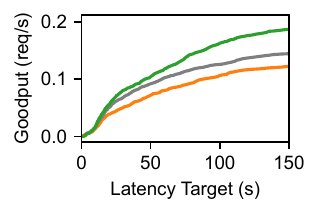}
  }%
  \subfloat[TIV$\to$A]{
    \includegraphics[width=0.46\columnwidth]{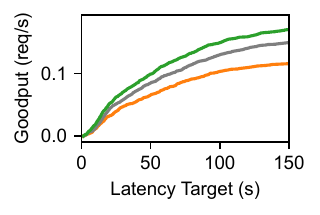}
  }

  \vspace{-0.5em}
  \subfloat[TIA$\to$A]{
    \includegraphics[width=0.46\columnwidth]{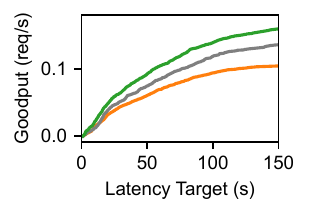}
  }%
  \subfloat[TIVA$\to$A]{
    \includegraphics[width=0.46\columnwidth]{figures/goodput_slo_16_gpu_omni_ra0.3_latencies/TIVA_A.pdf}
  }
  \vspace{-0.5em}
  \caption{Goodput of all 8 request types in Qwen 3 Omni vs.\ latency targets under 1/3 audio workload on 16 GPUs. (a)--(d): text output. (e)--(h): audio output.}
  \label{fig:appendix-omni-slo}
\end{figure}

Figure~\ref{fig:appendix-omni-slo} shows the goodput vs.\ latency target curves for all 8 request types (Section~\ref{sec:evaluation-omni} Figure~\ref{fig:evaluation-omni-slo-pertype} highlights TIVA$\to$T and TIVA$\to$A).
Cornfigurator achieves higher goodput compared to both baselines across all request types and latency targets.
The gap is especially pronounced for text-output types (top four), where Cornfigurator's colocation decisions reduce encoder interference and improve LLM throughput.
Audio-output types (bottom four) exhibit much longer latencies than text-output types due to the talker's iterative generation, and Cornfigurator's advantage comes from better allocation of talker--vocoder replicas.

\clearpage
\section{Qwen-Image Per-Type Goodput vs.\ Latency Target}\label{sec:appendix-qwen-image-slo}

\begin{figure}[t]
  \centering
  \includegraphics[width=0.85\columnwidth]{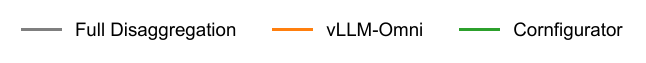}
  \vspace{-0.5em}

  \includegraphics[width=0.30\textwidth]{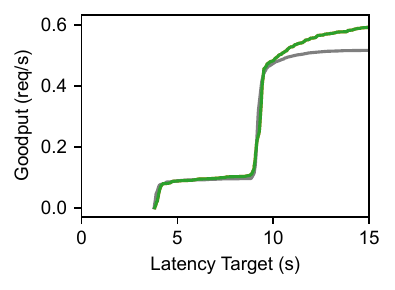}
  \vspace{-0.5em}
  \caption{Goodput of requests in Qwen-Image with text input and image output vs.\ latency targets on 16 GPUs.}
  \label{fig:appendix-qwen-image-slo}
\end{figure}

Figure~\ref{fig:appendix-qwen-image-slo} shows the goodput vs.\ latency target curve for Qwen-Image's single request type (text input, image output).
Cornfigurator overlaps with vLLM-Omni because the planner correctly identified vLLM-Omni's monolithic plan as optimal.
Full Disaggregation uses 2 LLM replicas + 7 SP=2 DiT replicas, as uniform sequence parallelism is enforced in this restricted search space.
The workload contains requests at 512$\times$512 and 1024$\times$1024 resolutions (weighted toward 1024), which have drastically different compute requirements.
The two visible inflection points in the curve correspond to these resolutions: the first bump around the 512$\times$512 compute time and the second around 1024$\times$1024.
At relaxed latency targets, the curves diverge because the monolithic deployment dedicates more GPUs to DiT replicas (no encoder GPU waste), achieving higher throughput.

\section{InternVL Workload Drift}\label{sec:appendix-workload-drift}

\begin{figure}[t]
  \centering
  \subfloat[Image probability 25\%]{
    \includegraphics[width=0.35\textwidth]{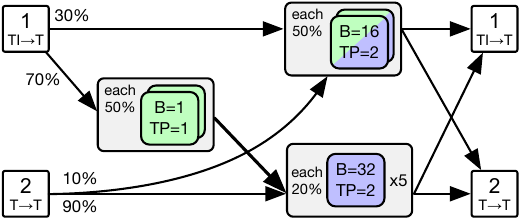}
  }

  \subfloat[Image probability 50\%]{
    \includegraphics[width=0.35\textwidth]{figures/eval-mllm-internvl-plan-0.50.pdf}
  }

  \subfloat[Image probability 75\%]{
    \includegraphics[width=0.35\textwidth]{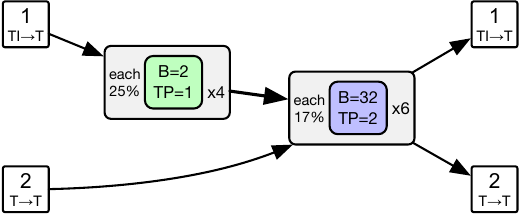}
  }
  \caption{Cornfigurator's deployment plans for InternVL 3 38B (\raisebox{-0.5ex}{\includegraphics[height=1.0em]{figures/eval-mllm-internvl-plan-legend.pdf}}) on 16 GPUs under workload drift in terms of image request type fraction.}\label{fig:appendix-internvl-drift}
\end{figure}

Figure~\ref{fig:appendix-internvl-drift} shows Cornfigurator's deployment plans for InternVL 3 38B on 16 GPUs under workload drift in terms of the fraction of image requests.
As image probability grows from 25\% to 75\%, the incoming workload is essentially being more image-heavy and exerts greater pressure on the image encoder.
This is why, when image probability increase from 25\% to 50\%, Cornfigurator's planner decides to additionally disaggregate the image encoder from the original monolithic executor.
As image probability further increases to 75\%, the image encoder becomes a more severe bottleneck of the system, and Cornfigurator responds by increasing the batch size of the image encoder to improve its throughput.
At this point, the LLM executor that used to be running with batch size 16 becomes the new bottleneck, so Cornfigurator also increases its batch size to 32---at which point the resulting plan is equivalent to pooling all encoders together, and all LLM executors together, resulting in the final plan.

\end{document}